%% file: ms.tex
\documentclass[conference]{IEEEtran}
\usepackage{listings}   
\usepackage{amsmath}
\usepackage[dvipsnames]{xcolor}
\usepackage{graphicx}
\usepackage{bm}
\newcommand{\rom}[1]{\uppercase\expandafter{\romannumeral #1\relax}}

\usepackage{pgfplots}
\pgfplotsset{width=10cm,compat=1.9}

\newcommand\fv[2]{$#1 = 0$ to $#2-1$}

\usepackage[caption=false]{subfig}

\usepackage{mathtools}
\DeclarePairedDelimiter\ceil{\lceil}{\rceil}

\usepackage{algorithm}
\usepackage[noend]{algpseudocode}

\makeatletter
\def\BState{\State\hskip-\ALG@thistlm}
\newcommand{\algmargin}{\the\ALG@thistlm}
\makeatother

\newlength{\whilewidth}
\algdef{SE}[parWHILE]{parWhile}{EndparWhile}[1]
  {\parbox[t]{\dimexpr\linewidth-\algmargin}{%
     \hangindent\whilewidth\strut\algorithmicwhile\ #1\ \algorithmicdo\strut}}{\algorithmicend\ \algorithmicwhile}%
\algnewcommand{\parState}[1]{\State%
  \parbox[t]{\dimexpr\linewidth-\algmargin}{\strut #1\strut}}
\newcommand\blfootnote[1]{%
  \begingroup
  \renewcommand\thefootnote{}\footnote{#1}%
  \addtocounter{footnote}{-1}%
  \endgroup
}

\usepackage[none]{hyphenat}

\ifCLASSINFOpdf
\else
\fi

\begin{document}
\title{A Design Methodology for Efficient Implementation of Deconvolutional Neural Networks on an FPGA}


% ## Author
% \author{\IEEEauthorblockN{Xinyu Zhang}
% \IEEEauthorblockA{University of California, San Diego\\
% xiz368@ucsd.edu}
% \and
% \IEEEauthorblockN{Homer Simpson}
% \IEEEauthorblockA{Twentieth Century Fox\\
% Email: homer@thesimpsons.com}
% \and
% \IEEEauthorblockN{Homer Simpson}
% \IEEEauthorblockA{Twentieth Century Fox\\
% Email: homer@thesimpsons.com}
% \and
% \IEEEauthorblockN{James Kirk\\ and Montgomery Scott}
% \IEEEauthorblockA{Starfleet Academy\\
% San Francisco, California 96678-2391\\
% Telephone: (800) 555--1212\\

\author{\IEEEauthorblockN{Xinyu Zhang,
Srinjoy Das, Ojash Neopane and
Ken Kreutz-Delgado}
\IEEEauthorblockA{University of California, San Diego\\
Email: 
xiz368@ucsd.edu,
s2das@ucsd.edu,
oneopane@ucsd.edu,
kreutz@ucsd.edu}}


%Telephone: (800) 555--1212, Fax: (888) 555--1212}


\maketitle

\input{content/0.abstract}


% \fi
\IEEEpeerreviewmaketitle

\input{content/1.Introduction}
\input{content/2.Background}
\input{content/3.Deconvolution_Hardware_Design}
\input{content/4.Top_Down_Optimization}

\input{content/5.Evaluation}
\input{content/6.Conclusion}


% \section*{Acknowledgment}




%\bibliographystyle{IEEEtran}


% \end{thebibliography}

\bibliographystyle{IEEEtran}

\bibliography{ref}
\end{document}

%% file: content/0.abstract.tex
\begin{abstract}
\boldmath
In recent years deep learning algorithms have shown extremely high performance on machine learning tasks such as image classification and speech recognition. In support of such applications, various FPGA accelerator architectures have been proposed for convolutional neural networks (CNNs) that enable high performance for classification tasks at lower power than CPU and GPU processors. However, to date, there has been little research on the use of FPGA implementations of deconvolutional neural networks (DCNNs). DCNNs, also known as generative CNNs, encode high-dimensional probability distributions and have been widely used for computer vision applications such as scene completion, scene segmentation, image creation, image denoising, and super-resolution imaging. We propose an FPGA architecture for deconvolutional networks built around an accelerator which effectively handles the complex memory access patterns needed to perform strided deconvolutions, and that supports convolution as well. We also develop a three-step design optimization method that systematically exploits statistical analysis, design space exploration and VLSI optimization. To verify our FPGA deconvolutional accelerator design methodology we train DCNNs offline on two representative datasets using the generative adversarial network method (GAN) run on Tensorflow, and then map these DCNNs to an FPGA DCNN-plus-accelerator implementation to perform generative inference on a Xilinx Zynq-7000 FPGA. Our DCNN implementation achieves a peak performance density of 0.012 GOPs/DSP.

% , which outperforms CNN FPGA accelerators described in the literature.
% \comment{The GOPS here is actually larger than the UCLA Paper by a considerable margin, because they are using floating number but we use short fixed point. I already calculate the GOPS number quite strictly. Another contributing factor is that it's easier to archive higher utilization rate in smaller board in my opinion.}
\end{abstract}


% , Generative Model, Generative Adversarial Network, Deconvolution, Acceleration

%% file: content/1.Introduction.tex
\section{Introduction}

Deep learning algorithms have shown extremely high performance on machine learning tasks. In particular, convolutional neural networks (CNNs) have become the state-of-the-art for applications like computer vision and audio recognition \cite{schmidhuber2015deep} \cite{lecun2015deep} \cite{Goodfellow-et-al-2016}. To address the increasing demand for applications that require running neural network algorithms in real time on embedded devices, various high performance hardware platforms for discriminative CNN implementations have been proposed, including the use of distributed GPUs or customized accelerators like FPGAs and ASICs \cite{chakradhar2010dynamically} \cite{chen2014dadiannao}. In particular, FPGA-based accelerators have been proposed  because they have lower latency and consume less power than GPUs while being more flexible and configurable than ASICs \cite{zhang2015optimizing} \cite{peemen2013memory}.

\begin{figure}[h] 
  \centering
  \includegraphics[width=0.45\textwidth]{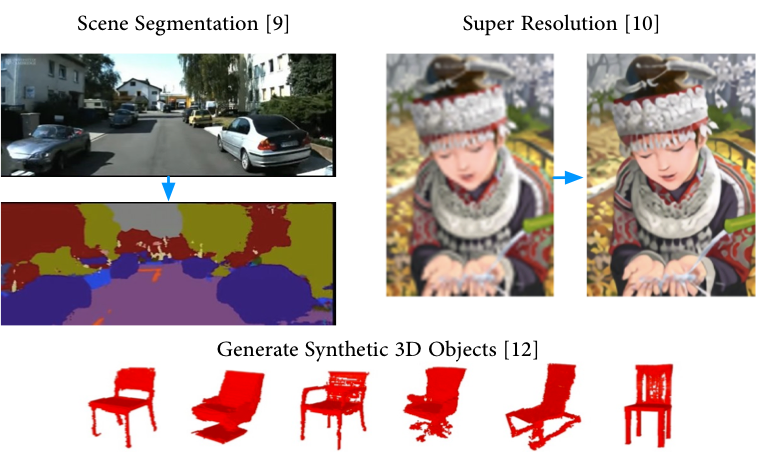}
  \caption{DCNNs work for pattern completion/generation (Images from \cite{3dgan} \cite{shi2016real} \cite{badrinarayanan2015segnet}).}
  \label{fig:gcnn_app}
\end{figure}

However, current FPGA accelerators focus on enhancing the performance of convolutional neural networks (CNNs), not deconvolutional neural networks (DCNNs). Unlike discriminative CNNs that effectively ``downsample'' the input to produce classification \cite{schmidhuber2015deep}, DCNNs are generative models capable of generating data by ``upsampling'' the input using deconvolution layers \cite{zeiler2010deconvolutional}. There are many applications of DCNNs, including multi-modal data modeling \cite{3dgan}, super resolution \cite{shi2016real} and image-to-image translation \cite{isola2016image} \cite{badrinarayanan2015segnet}  (see Fig. \ref{fig:gcnn_app}). Such applications motivate us to design an FPGA-based accelerator with the ability to execute deconvolution operations with high throughput and low cost.

%that could execute deconvolution layer with high throughput and low power for DCNNs. 

%  Imagine that a small and low-power imaging device installed on a vehicle could segment scenes in real time, Or supposed we could translate a live webcam video into a "cartoon" animation in real time.
% \vspace{-0.2cm}

% \vspace{-.1cm}
There are several issues that must be addressed to design an FPGA-based deconvolution accelerator. First, a direct translation of CPU-optimized deconvolution algorithms to an FPGA will generally lead to inefficient implementations. A suitable adaptation of the deconvolution operation to a hardware substrate such as FPGA is therefore necessary in order to achieve high performance with low implementation complexity. In addition, although recent research shows that discriminative CNNs are robust to low bitwidth quantization \cite{wu2016quantized} \cite{dundar1995effects}, it is important to be able to systematically study the effects of such bitwidth reductions on the quality of inference from a generative model such as DCNN implemented with finite precision on FPGA. Thus it is necessary to use metrics which quantify the effects of such approximations in DCNNs in order to achieve an efficient design optimized for performance and power.

To address the issues described above, we make the following contributions in this paper. 1) We create a deconvolution accelerator with reverse looping and stride hole skipping to efficiently implement deconvolution on an FPGA, where our proposed solution, in a nontrivial way, reuses the same computational architecture proposed for implementing a convolution accelerator in \cite{zhang2015optimizing}. 2) We propose a three-step procedure to design the deconvolution accelerator as follows. A) At the highest design level, we train DCNNs using the generative adversarial network method (GAN) \cite{goodfellow2014generative} and use statistical tests to quantitatively analyze the generative quality under different bitwidth precisions to select the most cost-efficient bitwidth. B) We use the roofline model proposed in \cite{zhang2015optimizing} to explore the design space in order to find the set of high-level constraints that achieves the best tradeoff between memory bandwidth and accelerator throughput.  C) We use loop unrolling and pipelining, memory partitioning, and register insertion to further optimize performance. 3) We validate our procedure via two implementations on a Xilinx Zynq-7000 FPGA.

The rest of this paper is organized as follows: Section \rom{2} provides background on the DCNN and the deconvolution layers. Section \rom{3} presents our methodology for efficiently implementing an FPGA-based deconvolution accelerator. Section  \rom{4} explains our three-step design methodology. Section \rom{5} shows our experimental results. Section \rom{6} concludes the paper.

%% file: content/2.Background.tex
% \vspace{-.25cm}
\section{Deconvolutional Neural Network}

% \subsection{Deconvolutional Neural Network}

\begin{figure}[h] 
  \centering
  \includegraphics[width=0.45\textwidth]{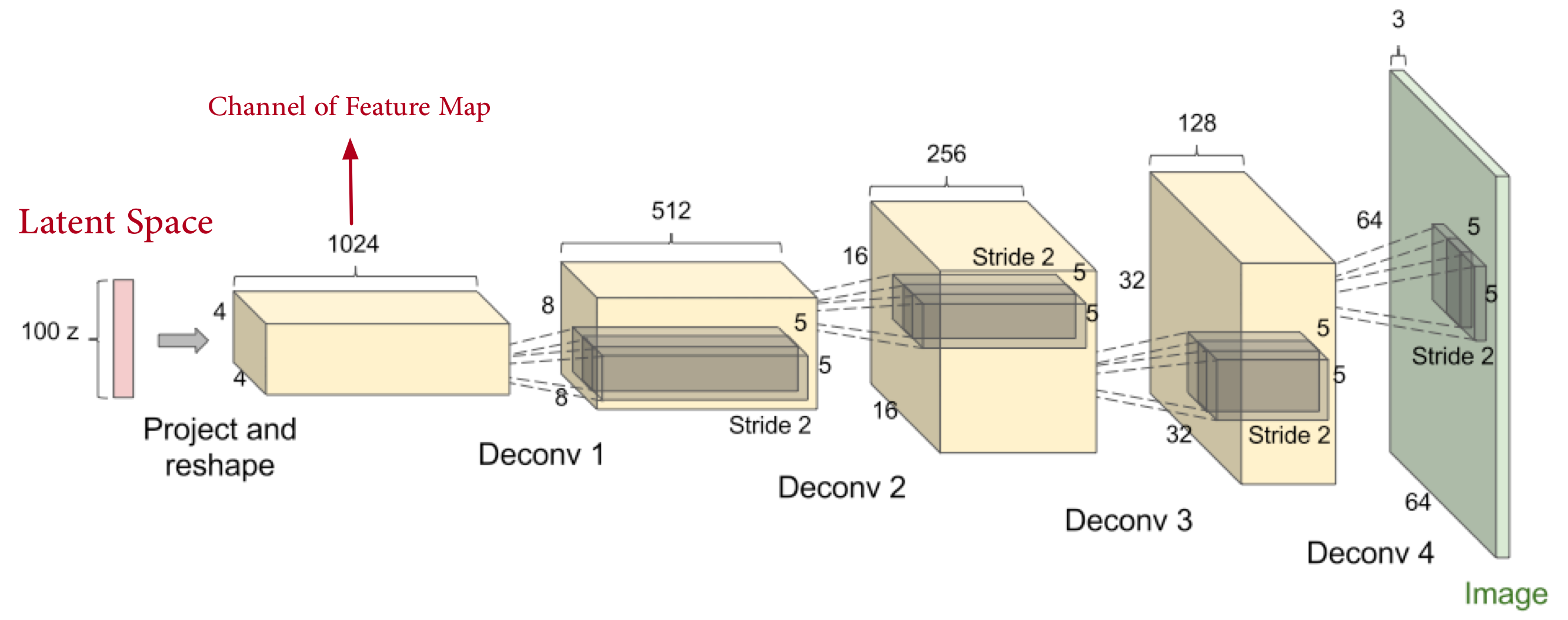}
  \caption{A DCNN that generates realistic 64x64 indoor scenes based on the use of four deconvolution layers that was trained on the Large-scale Scene Understanding (LSUN) Dataset \cite{radford2015unsupervised} \cite{yu2015lsun} (Image is taken and adapted from reference \cite{radford2015unsupervised}.)}
  \label{fig:dcnn_example}
\end{figure}

A deconvolutional neural network (DCNN) converts latent space representations to high-dimensional data similar to the training set by applying successive deconvolution operations in multiple layers \cite{noh2015learning_seg}. The latent space contains low-dimensional latent variables that provide a succinct (``conceptual'') representations of the possible outputs (e.g. an image). Thus a latent variable may correspond to ``chair'' with the associated output being the image of a chair ``generated'' by the DCNN (see Fig. \ref{fig:gcnn_app}). Fig. \ref{fig:dcnn_example} shows a 5-layer DCNN  developed in \cite{radford2015unsupervised} that consists of 4 deconvolutional layers. The first layer is fully-connected and transforms an input size of 1x100 to an output size of 1024x4x4; layers 2 to 5 are deconvolution layers that project low-dimensional feature maps into corresponding high-dimensional ones through successive layers.

% \subsection{Deconvolution Layer} \label{subsec:deconvolution_layer}

\begin{figure}[h] 
  \centering
  \includegraphics[width=0.47\textwidth]{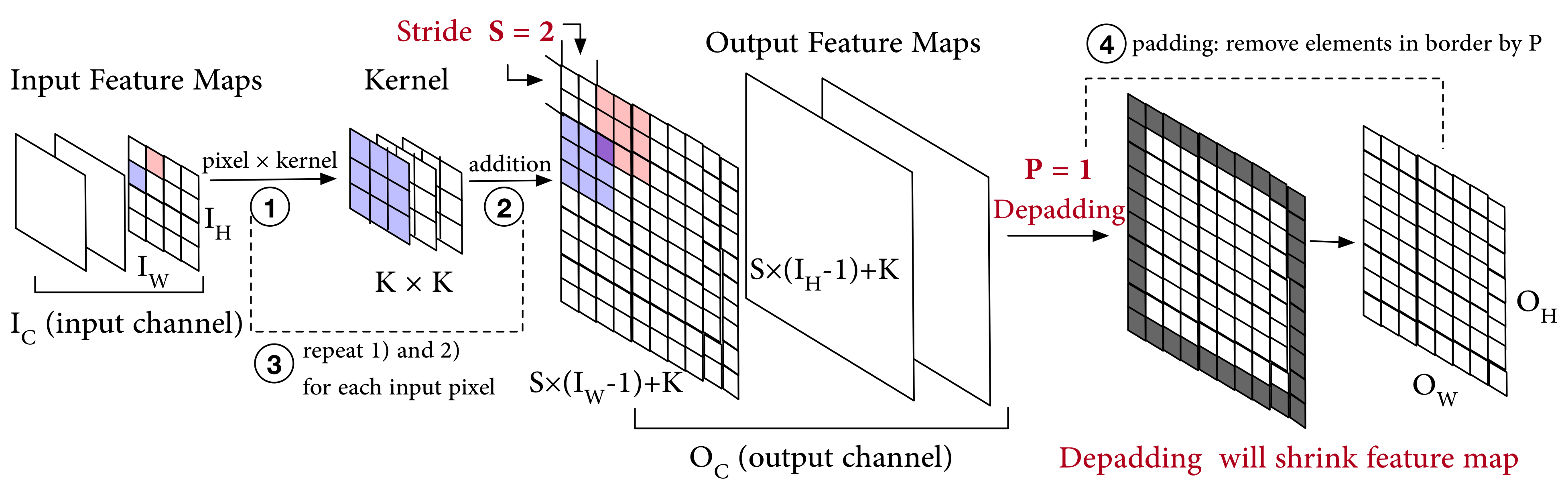}
  \caption{Visualization of a Single Deconvolution Layer. The four steps required to implement the deconvolutional layer are: (1) multiply a single input pixel $i_h,i_w$ by a $K\times K$ kernel; (2) add the result of step 1 to a local area in the output feature map that starts at $i_h\times S, i_w \times S$; (3) repeat 1 and 2 for all input pixels; (4) remove elements from output feature maps in the border by zero padding of size $P$.}
  \label{fig:deconv_vis}
\end{figure}

Fig. \ref{fig:deconv_vis} shows how a typical deconvolution layer works, where $S$ and $P$ denote the chosen values of stride and padding respectively for a given layer. The pseudo code of a deconvolution layer as implemented in CPU is shown in Algorithm. \ref{alg:deconv_cpu} which uses the loop variables defined in Fig. 4.

% \vspace{-0.75cm}
\blfootnote{By convention we use capital letters e.g. $O_H$ to denote specific parameters of the DCNN whereas small letters e.g. $o_h$ to denote its corresponding loop variable.}

\begin{figure}[h] 
  \centering
  \includegraphics[width=0.47\textwidth]{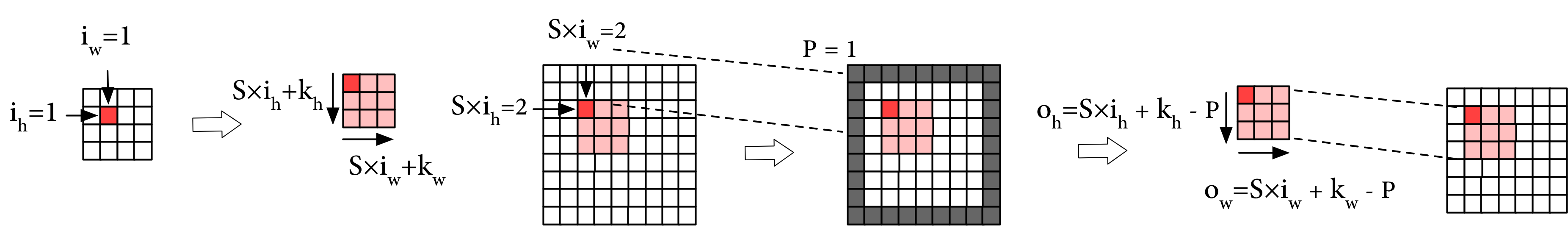}
  \caption{Visualization of Algorithm \ref{alg:deconv_cpu} with loop variables.}
  \label{fig:line8}
\end{figure}

\begin{algorithm}
\caption{Deconvolution in CPU}\label{alg:deconv_cpu}
\begin{algorithmic}[1]
\Procedure{Deconvolution}{}
\For{\fv{i_c}{I_C}}
    \For{\fv{i_h}{I_H}} 
        \For{\fv{i_w}{I_W}}
            \For{\fv{o_c}{O_C}}
                \For{\fv{k_h}{K}}
                    \For{\fv{k_w}{K}}
                        \State $o_h \gets S\times i_h + k_h - P$ 
                        \State $o_w \gets S\times i_w + k_w - P$
                        \State \parbox[t]{.6\linewidth}{%
    $\text{out}[o_c][o_h][o_w] \gets (\text{in}[i_c][i_h][i_w]$ \\
    $\times\, \text{kernel}[o_c][i_c][k_h][k_w])$}              
                    \EndFor
                \EndFor
            \EndFor
        \EndFor
    \EndFor
\EndFor
\EndProcedure
\end{algorithmic}
\end{algorithm}
\setlength{\textfloatsep}{0pt}% Remove \textfloatsep
The relation of the input size $I_H\times I_W$ to output size $O_H \times O_W$ after applying stride and padding are given in the following equations \cite{dumoulin2016guide}: 
    \begin{align}
    \begin{split} 
    O_H = S\times(I_H  - 1) + K - 2P \\
    O_W = S\times(I_W  - 1) + K - 2P
    \end{split}
    \label{eq:feature_map_size}
    \end{align}

%% file: content/3.Deconvolution_Hardware_Design.tex
\section{Deconvolution Hardware Design}

\begin{figure}[h] 
      \centering
      \includegraphics[width=0.48\textwidth]{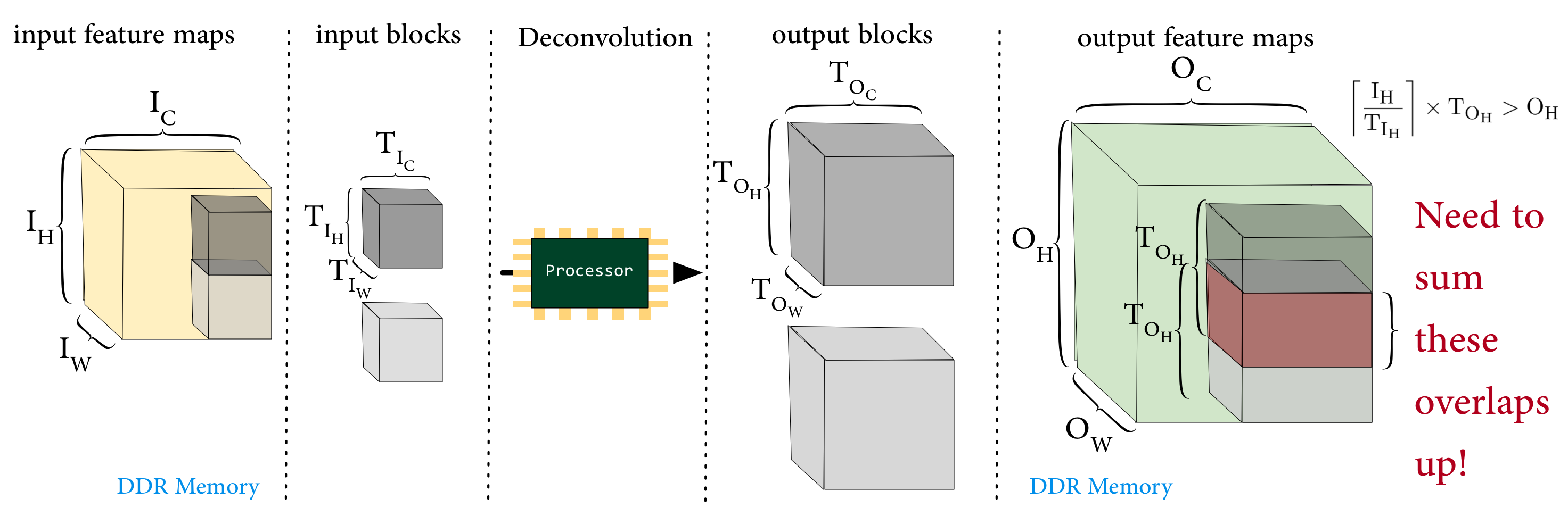}
      \caption{Traditional implementation of deconvolution. The input feature map is first divided into separate blocks and PEs read each block from DDR and process the deconvolution operations on this block. Finally the results are stored back to the DDR. This procedure is inefficient and can be circumvented as described in the text. }
      \label{fig:tile_demo}
    \end{figure}

An FPGA accelerator usually consists of processing elements (PEs), registers, and local memory elements referred to as block RAMs (BRAMs). Processing elements operate on data provided by the local memory, which communicates with external dual data rate (DDR) memory using direct memory access (DMA). Fig. \ref{fig:tile_demo} shows a traditional implementation of deconvolution, where $T_{I_H}$, $T_{I_W}$, $T_{I_C}$, $T_{O_H}$, $T_{O_W}$, and $T_{O_C}$ are the dimensions of the input and output block. Replacing $I_H$, $O_H$ with $T_{I_H}$, $T_{O_H}$ in Eq. \ref{eq:feature_map_size}, we have:
    \begin{equation} 
    T_{O_H} = S \times (T_{I_H} - 1) + K - 2P
    \end{equation}
Here the zero padding $P=0$ because blocks are inside input feature maps. However, Eq. \ref{eq:t_oh_oh} shows that deconvolution results of input blocks overlap with each other:
    \begin{equation} 
    \ceil*{\frac{I_H}{T_{I_H}}} \times T_{O_H} > O_H
    \label{eq:t_oh_oh}
    \end{equation}
    
Deconvolution arithmetic requires overlapping regions between output blocks to be summed together \cite{dumoulin2016guide} which can be realized in processor-based implementations. However handling such operations in FPGAs requires either the design of additional hardware blocks which creates overhead or communicating with a host processor which can increase system latencies thereby precluding real-time applications.
    
% Traditionally, the overlapping regions between output blocks (shown in Fig. \ref{fig:tile_demo}) are summed together \cite{dumoulin2016guide}. However, for FPGAs, handling these additional summation operations requires either designing a \textbf{hardware} finite state machine or writing a CPU program; both of which would eliminate the ability for real-time applications.

\subsection{Reverse Looping} \label{subsec:reverse_loop}

\begin{figure}[h] 
  \centering
  \includegraphics[width=0.3\textwidth]{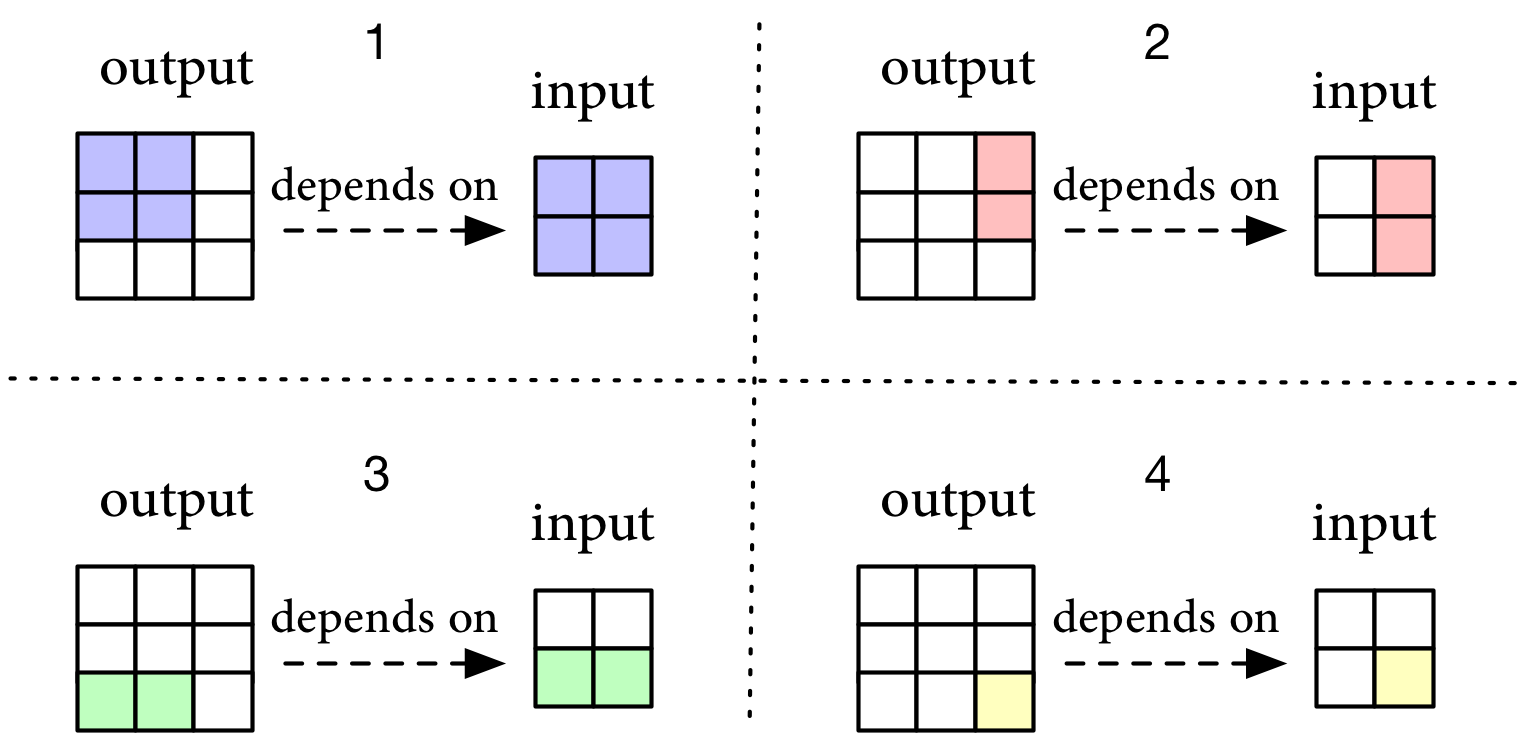}
  \caption{An efficient way to deconvolve. We first take a block in the output space and determine which inputs are needed to calculate the values in the block.  Then, for each block, the input is deconvolved and the appropriate output is extracted.  This is done sequentially until values have been computed for the entire output space.}
  \label{fig:reverse_loop}
\end{figure}

To avoid the overlapping sum problem, we propose a technique called reverse looping, where instead of directly deconvolving the input space, we use the output space to determine which input blocks to deconvolve and thus eliminating the need for the additional summation operations described above. This procedure is indicated in Fig. \ref{fig:reverse_loop}.

% This guarantees that there is only a single write to a pixel in an output feature map during deconvolution. 

The loop iterations over $i_h$ and $i_w$ in the CPU implementation shown in Algorithm \ref{alg:deconv_cpu} need to be recast over $o_h$ and $o_w$. Referring to Algorithm \ref{alg:deconv_cpu} and Fig. \ref{fig:line8}, we have:
    \begin{equation} 
    o_h = i_h \times S + k_h - P
    \label{eq:oh_up_ih_S_hk}
    \end{equation}
Rearranging terms, we get:
    \begin{equation} 
    i_h = \frac{o_h+P-k_h}{S}
    \label{eq:oh_P_hk_S}
    \end{equation}
Unfortunately Eq. \ref{eq:oh_P_hk_S} generally results in a non-integer value for the loop variable $i_h$, which is invalid \cite{dumoulin2016guide}. One way to address this problem would be to monitor $i_h$ so that fractional values can be discarded. However this would consume additional hardware resources and create unnecessary latencies in the system. 

% To solve this issue, we propose a technique called stride hole skipping which allows us to manipulate $o_h$ to ensure Eq. \ref{eq:oh_P_hk_S} to be integer.

\subsection{Stride Hole Skipping}

In this section, we propose a technique called stride hole skipping to ensure $i_h$  of Eq. \ref{eq:oh_P_hk_S} is an integer.  Toward this end, we recast $o_h$ in terms of two new variables, $o_h'$ and $f_h$ and show that this leads to an effective way of solving the aforementioned problem. First note that a sufficient condition for $i_h$ to be an integer in Eq. \ref{eq:oh_P_hk_S} is: 
    \begin{equation} 
    (o_h+P-k_h) \bmod S = 0
    \label{eq:oh_mod}
    \end{equation}
Assuming $\frac{O_H}{S}$ is an integer ($O_H$ is defined in Eq. \ref{eq:feature_map_size}), we can recast $o_h$ as follows:
    \begin{align}
    \begin{split} 
    o_h = S\times o_h' + f_h,\quad f_h \in \{0,1,...,S-1\} \\
    o_h' \in \{0,1,..., \frac{O_H}{S}-1\}
    \end{split}
    \label{eq:oh_rewrite}
    \end{align}
Using the definition of $o_h$ in Eq. \ref{eq:oh_mod}, we can recast the sufficient condition Eq. \ref{eq:oh_mod} in terms of $f_h$ as below:
    \begin{equation} 
    (f_h + P - k_h) \bmod S = 0
    \label{eq:oh_mod_2}
    \end{equation}
Eq. \ref{eq:oh_rewrite} implies that we can rewrite $f_h$ as:
    \begin{equation} 
    f_h = S - ((P-k_h) \bmod S)
    \label{eq:offset_22}
    \end{equation}
This can be verified by plugging in  Eq. \ref{eq:offset_22} into Eq. \ref{eq:oh_mod_2} which yields the following identity:
    \begin{equation} 
    (P - k_h - (P - k_h) \bmod S) \bmod S = 0
    \label{eq:p_hk}
    \end{equation}

To prevent $f_h$ from taking a value equal to $S$, we enforce the additional condition:
    \begin{equation} 
    f_h = (S - ((P-k_h) \bmod S)) \bmod S
    \label{eq:offset_2}
    \end{equation}
By using Eq. \ref{eq:offset_2} to choose values for $f_h$,  we can ensure that $o_h$  computed from Eq. \ref{eq:oh_rewrite} meets the condition in Eq. \ref{eq:oh_mod}. Therefore we can avoid the previously mentioned issue of discarding fractional values of $i_h$ that we would otherwise encounter from a direct application of Eq. \ref{eq:oh_P_hk_S}. The pseudo code for deconvolution on FPGA is shown in Algorithm \ref{alg:proposed_deconv}.

\begin{algorithm}
\caption{Our FPGA Implementation of Deconvolution}\label{alg:proposed_deconv}
\begin{algorithmic}[1]
\Procedure{ReverseDeconvolution}{}
\For{\fv{k_h}{K}}
    \For{\fv{k_w}{K}}
        \For{\fv{o_h'}{\frac{T_{O_H}}{S}}}
            \For{\fv{o_w'}{\frac{T_{O_W}}{S}}} \Comment loop $T_{O_W}$
                \For{\fv{o_c}{T_{O_C}}}  \Comment loop $T_{O_C}$
                    \For{\fv{i_c}{T_{I_C}}}\Comment loop $T_{I_C}$
                        \State \small{COMPUTE}$(k_h, k_w, o_h', o_w', o_c, i_c)$
                    \EndFor
                \EndFor
            \EndFor
        \EndFor
    \EndFor
\EndFor
\EndProcedure
\Procedure{Compute}{$k_h, k_w, o_h', o_w', o_c, i_c$}
\State $f_h \gets (S - ((P - k_h) \bmod S)) \bmod S$
\State $f_w \gets (S - ((P - k_w) \bmod S)) \bmod S$
\State $o_h = o_h'\times S + P + f_h$
\State $o_w = o_w'\times S + P + f_w$
\State $i_h \gets {(o_h-k_h)}/{S}$
\State $i_w \gets {(o_w-k_w)}/{S}$
\State $\text{out}[o_c][o_h][o_w] \gets \text{in}[i_c][i_h][i_w]\times\text{kernel}[o_c][i_c][k_h][k_w]$
\EndProcedure
\end{algorithmic}
\end{algorithm}

%The difference in architecture between the deconvolution accelerator in Algorithm \ref{alg:proposed_deconv} and the convolution accelerator proposed in \cite{zhang2015optimizing} is the local memory addressing (shown in red). The close similarity means we can share the basic computation architecture for both convolution and deconvolution layers. 

% \subsection{Input Cache Inference}

%% file: content/4.Top_Down_Optimization.tex
% \vspace{-.3cm}
\section{Three-Step Design Methodology}
\subsection{Statistical Analysis}

% 8 bit of precision is usually enough for accurate performance of discriminative neural networks \cite{vanhoucke2011improving}, but this notion does not generally hold true for generative neural networks \cite{neopane2016nonparametric}. 

It is important to study the effect of bitwidth reduction on the quality of inference from the generative model. To find out the most cost-efficient bitwidth for DCNNs, we fix $T_{O_H}, T_{O_W}, T_{O_C}, T_{I_C}$, and study the trade-off between generative quality and implementation complexity over a range of bitwidths using statistical analysis. Quantifying generative models using traditional techniques such as  Kullback-Leibler divergence and log-likelihood are not feasible in high-dimensional settings such as the typical setting deconvolutional neural networks are used in. To overcome this drawback, we apply nonparametric goodness of fit testing.  Specifically, we apply the Relative Maximum Mean Discrepancy (RMMD) Test proposed by \cite{bounliphone2015test_rmmd} to measure and compare the performance of our system at different bitwidths.

The RMMD is an extension of the Maximum Mean Discrepancy (MMD) two sample test proposed by \cite{gretton2012kernelmmd}.  Given samples $\{X_i\}_{i=1}^{m}$ and $\{Y_i\}_{i=1}^{n}$ from distributions $P_x$ and $P_y$ the MMD test statistic is given by: $$\text{MMD}^2(X,Y) =  \frac{1}{m(m-1)}\sum_{i=1}^m\sum_{j\neq i}^m k(x_i,x_j)$$ $$+ \frac{1}{n(n-1)}\sum_{i=1}^n\sum_{j\neq i}^n k(y_i,y_j) - \frac{2}{mn}\sum_{i=1}^m\sum_{j=1}^n k(x_i,y_j)$$ the null hypothesis $H_0: P_x = P_y$ is tested versus alternative $H_1: P_x \neq P_y$.
In the above equation, k is the Radial Basis Function given by $$k(x,y) = \exp{||x - y||}$$

The RMMD test builds upon the standard MMD framework by computing the MMD test statistic between two pairs of distributions.  Given samples $\{X_i\}_{i=1}^{m}$, $\{Y_i\}_{i=1}^{n}$, and $\{Z_i\}_{i=1}^{r}$ respectively from the training data, low-bitwidth DCNN, and full-precision DCNN, RMMD tests the null hypothesis $H_0:MMD^2(X,Y)<MMD^2(X,Z)$ against the alternative $H_1:MMD^2(X,Z)<MMD^2(X,Y)$. \cite{bounliphone2015test_rmmd} shows that the p-values for testing $H_0$ against $H_1$ are given by: $$p \leq \Phi(- \frac{\text{MMD}_u^2(X_m,Y_n) - \text{MMD}_u^2(X_m,Z_r)}{\sqrt{\sigma_{XY}^2 + \sigma_{XZ}^2 - 2\sigma_{XYXZ}}})$$
where $\Phi$ is the Normal Cumulative Distribution Function.  The p-value in the above equation indicates the probability that, based on the observed samples, the distribution based on the low bitwidth DCNN is closer to the training data than the distribution based on the full precision DCNN is to the training data.  Using this interpretation: 

%FROM Ojash
\begin{itemize}
     \item a p-value $> 0.5$ indicates the low bitwidth DCNN is more similar to the training data
     \item a p-value $< 0.5$ indicates the full precision DCNN is more similar to the training data
\end{itemize}

%% NEWLY ADDED After Ojash Text
% To study the trade-off among bitwidth, generative quality, and system complexity, we set bitwidth as the $X$ axis and $\text{p-value} \times \text{slack}$ or $\text{p-value}/\text{power}$ as the $Y$ axis, where the slack denotes the worst slack. The two curves are shown in Fig. \ref{fig:curve_bit}. Both curves peak at bitwidth 12, which we take to be a good choice because it represents a high p-value (generative quality) with a low power consumption and large worst slack.

% \begin{figure}[h]
% \centering
% \begin{tabular}{cc}
% \subfloat[p-value$/$power vs bitwidth]{\includegraphics[width = 0.225\textwidth]{images/curve_power.png}} &
% \subfloat[p-value$\times$slack vs bitwidth]{\includegraphics[width = 0.225\textwidth]{images/curve_slack.png}} \\ 
% \end{tabular}
% \caption{Approximate concave curves based on trade-off among bitwidth, quality and system complexity}
% \label{fig:curve_bit}
% \end{figure}

\subsection{Roofline Analysis}  \label{subsec:dsp}

\begin{figure}[h] 
  \centering
  \includegraphics[width=0.4\textwidth]{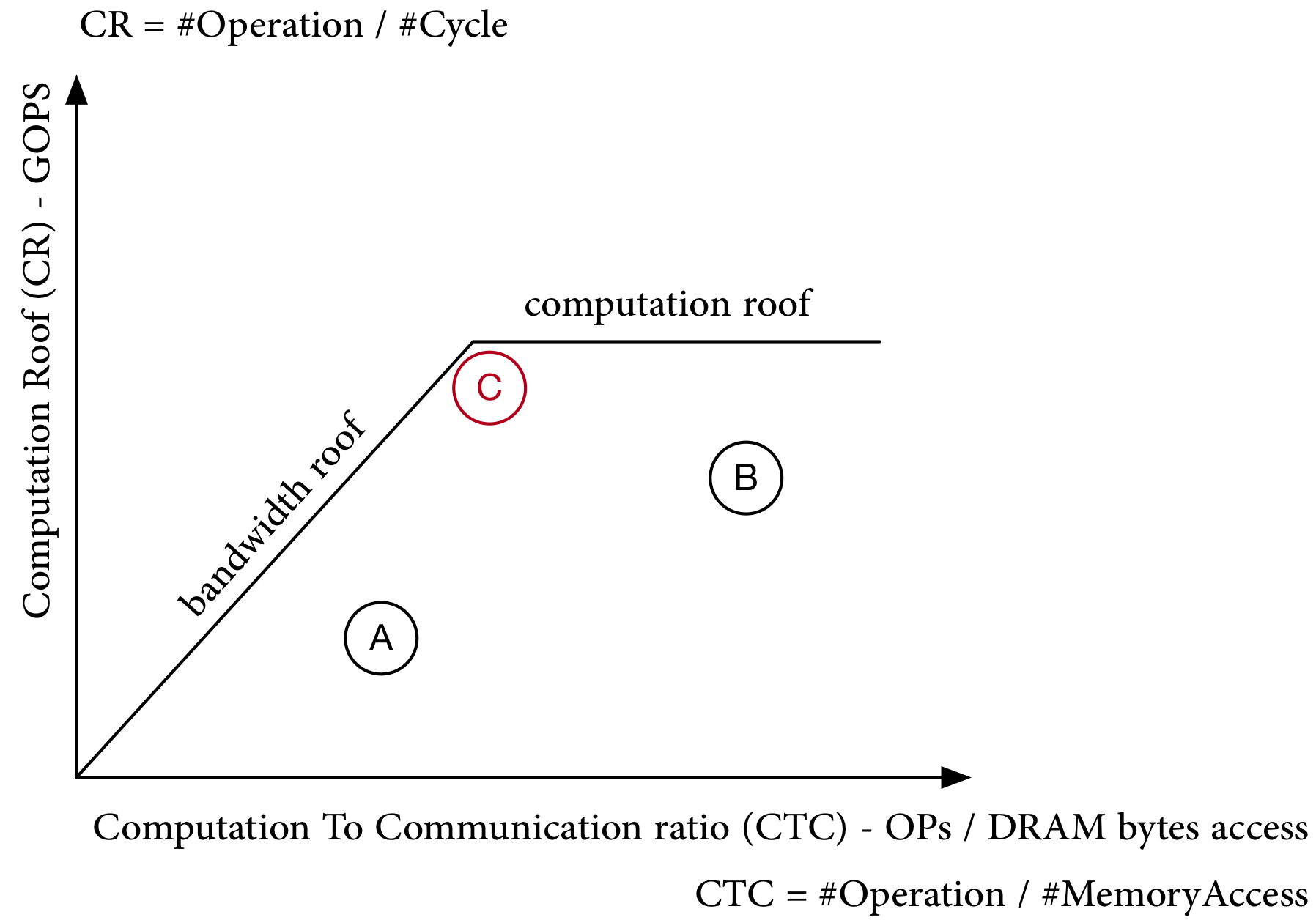}
  \caption{Roofline Model, adopted from \cite{zhang2015optimizing}}
  \label{fig:roofline}
\end{figure}

The generative quality is determined by choosing the optimal bitwidth using the previously described procedure. Following this we turn to further increasing the throughput by optimizing with respect to $T_{O_H}$, $T_{O_W}$, $T_{O_C}$, and $T_{I_C}$, which are the height, width, channel size of output block, and channel size of input block respectively (see Fig. \ref{fig:tile_demo}). This is done using roofline analysis \cite{zhang2015optimizing}. Fig. \ref{fig:roofline} shows an example roofline plot where the $X$ axis denotes the number of operations per memory access and $Y$ axis denotes the number of operations per cycle.

In this drawing, A, B and C correspond to designs of accelerator with different values of $T_{O_H}, T_{O_W}, T_{O_C}, T_{I_C}$. Design A transfers too much data, so computation speed is low, and therefore falls well beneath the computation roof. Design B lies well beneath the bandwidth roof, which means the system performance is dominated by memory transfers. Design C is more efficient than A and B with its balance between computation speed and memory bandwidth. This technique is described in \cite{zhang2015optimizing} and is used for the design of convolution accelerator. We apply roofline analysis to design deconvolution accelerator and estimate the computation to communication ratio (CTC) and computational roof (CR) for a given layer.

% We use roofline analysis to estimate the number of operations, memory accesses, and computation cycles for a given layer.

% \subsubsection{Number of Operations}

% For an accelerator architecture that implements the Algorithm \ref{alg:deconv_cpu}, the number of operations of a deconvolution layer is given by:
% \begin{equation}
% \#\text{OP} = I_C \times O_C \times I_H \times I_W \times K^2\times6 
% \end{equation}
% which includes all addressing, multiply and addition operations of deconvolution.

\subsubsection{Computation to Communication Ratio}
Let $\alpha_{in}$, $\alpha_{w}$, $\alpha_{out}$ and $B_{in}$, $B_{w}$, $B_{out}$ denote the trip counts and buffer sizes of memory accesses to input/output feature maps, weights, respectively. The CTC is given by:
    \begin{align}
    \begin{split} 
    \text{CTC} = \frac{\text{total number of operations}}{\text{total amount of external memory access}} \\
    = \frac{2 \times I_C \times O_C \times I_H \times I_W \times K^2}{\alpha_{in} B_{in} + \alpha_{w} B_{w} + \alpha_{out} B_{out}}
    \end{split}
    \label{eq:num_mem_acc}
    \end{align}
    \begin{equation}
    \alpha_{out} = \frac{O_C}{T_{O_C}} \frac{O_H}{T_{O_H}}, \alpha_{in} = \alpha_{w} = \frac{I_C}{T_{I_C}} \alpha_{out}
    \end{equation}
    \begin{equation}
    B_{in} = T_{I_C} \left(\frac{T_{O_H} + K}{S}\right) \left(\frac{T_{O_W} + K}{S}\right)
    \end{equation}
    \begin{equation}
    B_{out} = T_{O_C} T_{O_H} T_{O_W},\qquad B_{weight} = T_{O_C} T_{I_C} K^2
    \end{equation}
    \begin{equation}
    0 \leq B_{in} + B_{w} + B_{out} \leq \text{BRAM}_{\text{capacity}}
    \end{equation}

\subsubsection{Computation Roof}

Let $\text{PD}$ denotes the pipeline depth and $\text{II}$ is the number of cycles between the start of each loop iteration $T_{O_W}$, the CR is given by:
    \begin{align}
    \begin{split}
    \text{CR} = \frac{\text{total number of operations}}{\text{number of execution cycles}} \\
    =  \frac{2 \times I_C \times O_C \times I_H \times I_W \times K^2}{\alpha_{in} K^2 T_{O_H} (\text{PD} + \text{II}(T_{O_W} - 1))}
    \end{split} 
    \label{eq:num_cc}
    \end{align}
where 
    \[
    \begin{cases}
    0 \leq T_{O_C} T_{I_C} \leq (\#\,\text{of\,DSPs}) \\
    0 < T_{I_C} \leq I_C \\
    0 < T_{O_C} \leq O_C \\
    0 < T_{O_H} \leq O_H \\
    0 < T_{O_W} \leq O_W
    \end{cases}
    \]

Note that $0 \leq T_{O_C} T_{I_C} \leq (\#\,\text{of\,DSPs})$ will not hold true when the bitwidth is greater than 18, because the maximum bitwidth of the  multipliers used in our implementation is 18-bit \cite{xilinx:ug479}. Since we use a bitwidth of 12 in all our experiments this constraint is therefore valid.

\subsection{VLSI Level Optimization} \label{subsec:vlsi_opt}

\subsubsection{Loop Unrolling and Pipelining}

\begin{figure}[h] 
  \centering
  \includegraphics[width=0.40\textwidth]{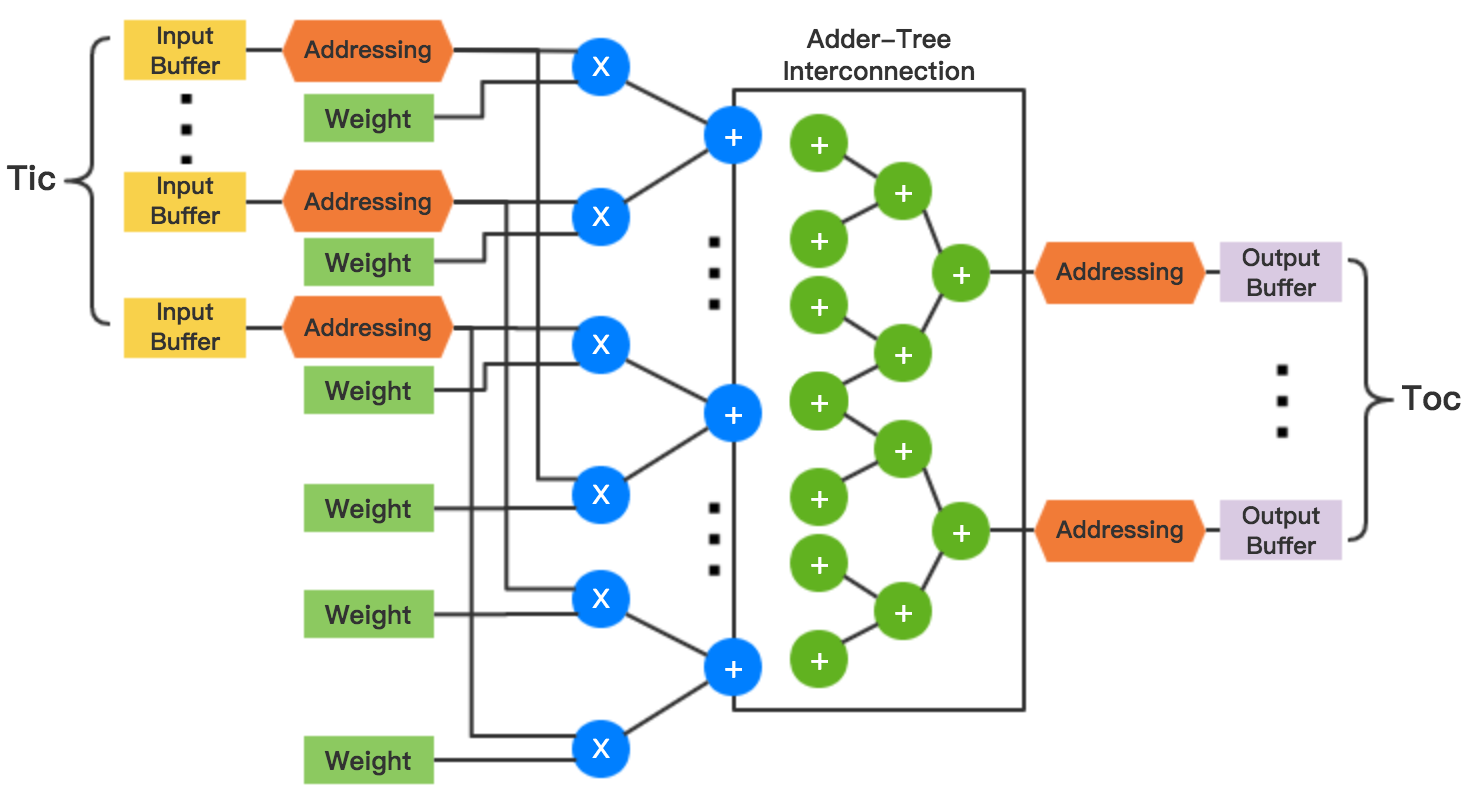}
  \caption{Processing Engine}
  \label{fig:processing_element}
\end{figure}

Loop unrolling is a key technique of high level synthesis \cite{Coussy:2008:HSA:1457713_hls}. It works by generating parallel hardware to accelerate FPGA program execution. The innermost loop $T_{O_C}$ and $T_{I_C}$ in Algorithm \ref{alg:proposed_deconv} are unrolled and can be executed in a constant amount of cycles $P$, which forms the processing engine as shown in Fig. \ref{fig:processing_element}. We also pipeline the loop $T_{O_W}$ with carried dependency of 2.

\subsubsection{Register Insertion}

\begin{figure}[h] 
  \centering
  \includegraphics[width=0.35\textwidth]{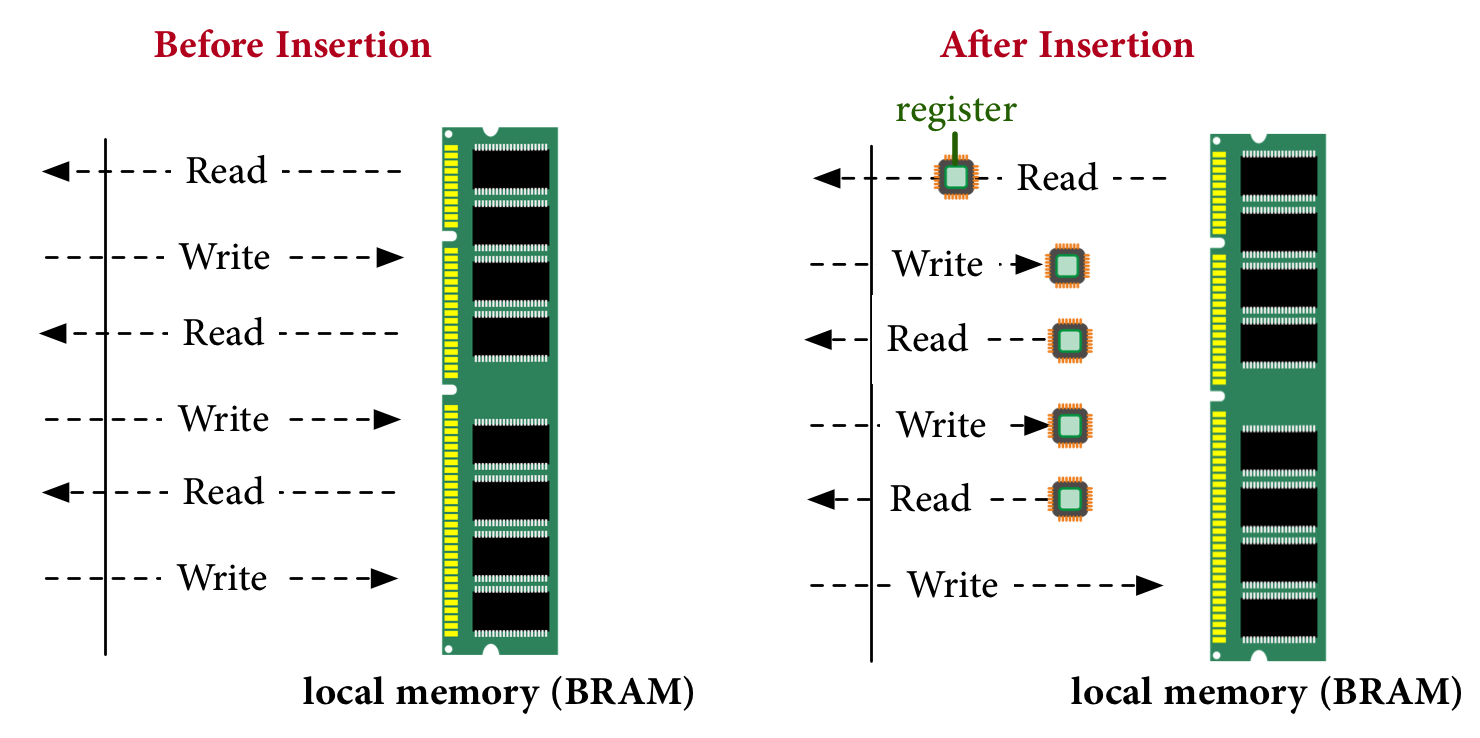}
  \caption{Insert register to reduce local memory (BRAM) writes}
  \label{fig:register_insertion}
\end{figure}

The critical path length and pipeline interval are constrained by the on-chip local memory bandwidth, especially when the size of the processing engine is large.  To further improve performance, we insert registers to economize local memory bandwidth, which is illustrated in Fig. \ref{fig:register_insertion}.

%% file: content/5.Evaluation.tex
% \vspace{-.3cm}
\section{Evaluation}

\subsection{Statistical Analysis}

\begin{figure}[h]
\centering
\begin{tabular}{cc}
\subfloat[p-value$/$power vs bitwidth]{\includegraphics[width = 0.225\textwidth]{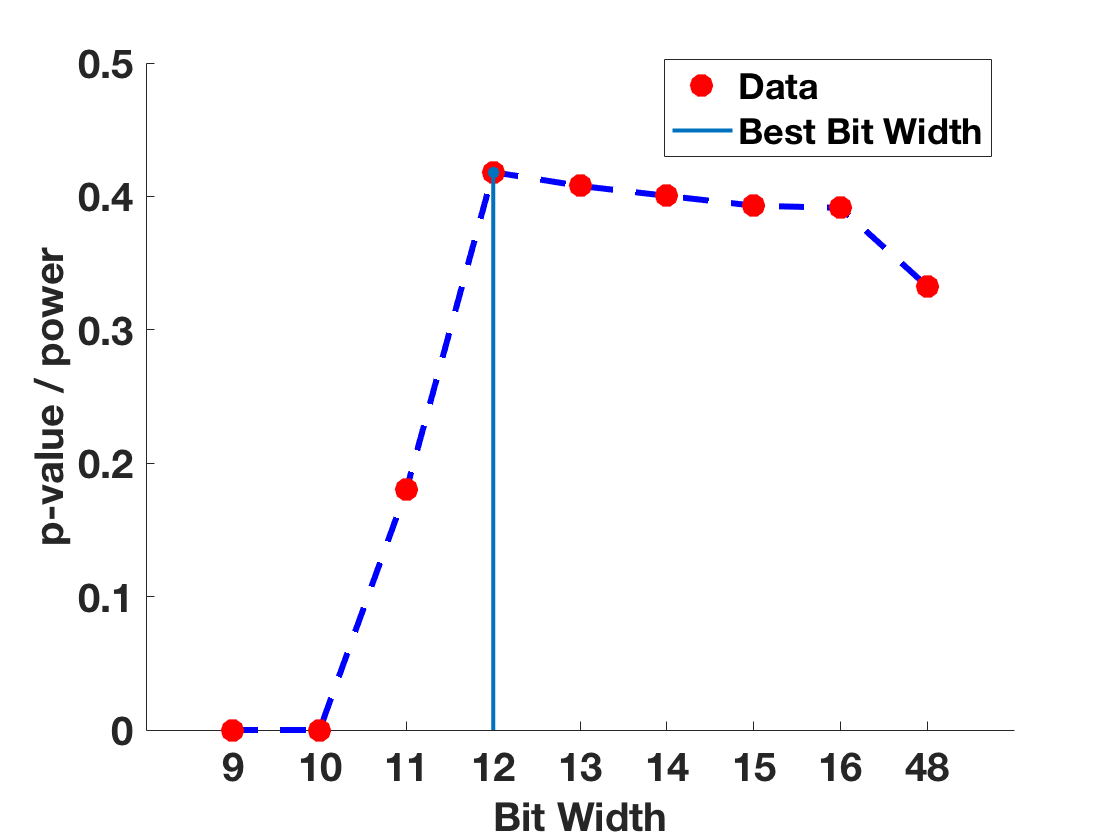}} &
\subfloat[p-value$\times$slack vs bitwidth]{\includegraphics[width = 0.225\textwidth]{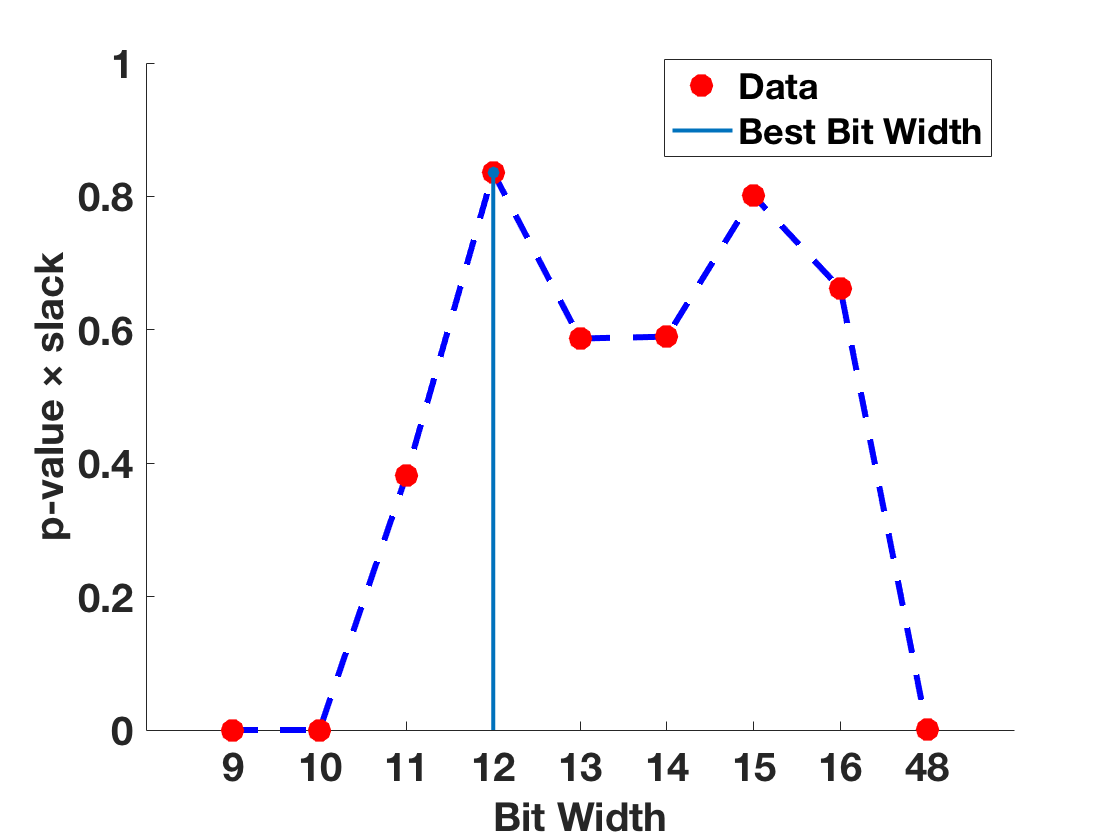}} \\ 
\end{tabular}
\caption{Approximate concave curves based on trade-off between generative quality and implementation complexity.}
\label{fig:curve_bit}
\end{figure}

Previous work such as that described in \cite{neopane2016nonparametric} has shown the effectiveness of using high-dimensional nonparametric tests to determine optimal parameters for generative inference in hardware. For designing the deconvolution accelerator we follow a similar approach and use the RMMD test framework outlined in Section IV A to choose the optimal bitwidth for our system. For this purpose, we trained two DCNNs through the method described in \cite{radford2015unsupervised} on the MNIST and CelebA Human Face datasets \cite{liu2015faceattributes}. To study the trade-off between generative quality and system complexity over a range of bitwidths, we determine $\text{p-value} \times \text{minimum slack}$ and $\text{p-value} / \text{power}$ as a function of bitwidths. The two curves are shown in Fig. \ref{fig:curve_bit}. Both curves peak at bitwidth 12, which we take to be a good choice because it represents a high p-value (generative quality) with a low power consumption and high minimum slack.

\subsection{Hardware System}

\begin{figure}[h] 
  \centering
  \includegraphics[width=0.45\textwidth]{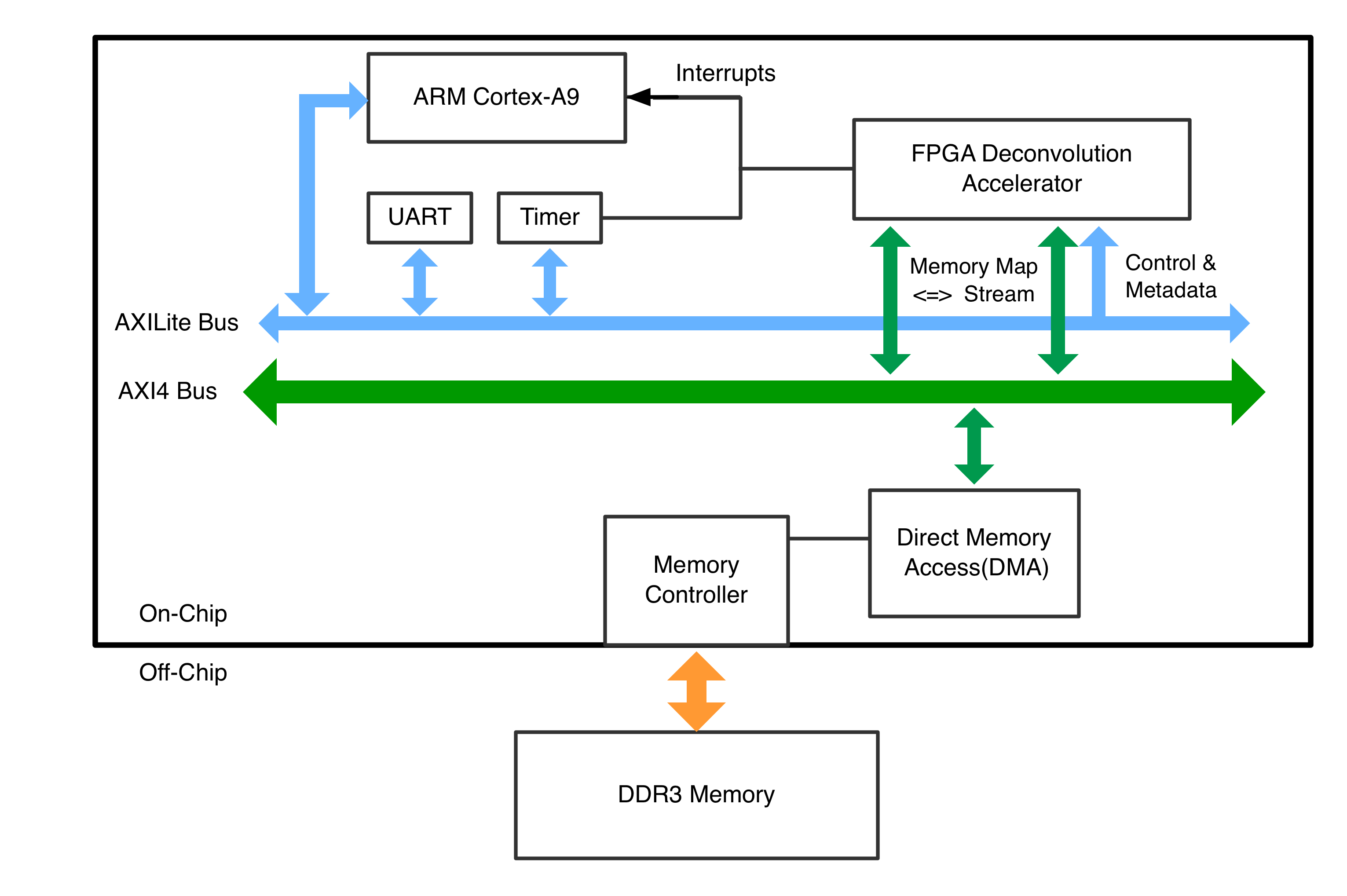}
  \caption{Overview of Implementation Block Diagram.}
  \label{fig:overview_block}
\end{figure}

We implemented the deconvolution accelerator IP with Vivado HLS (v2016.2). We use \textit{ap\_fixed.h} from Vivado Math Library to implement fixed point arithmetic operations with arbitrary bitwidth precision, and use \textit{hls\_stream.h \& ap\_axi\_sdata.h} to model streaming data structure. The hardware system is built on a Zynq-7000 FPGA XZ7020 with Vivado Design Suite and Xilinx SDK. The FPGA 7Z020 is programed with our accelerator IP and the ARM processor is used to initialize the accelerator, set parameters, and transfer data for each layer. An overview of the implementation block diagram is in Fig. \ref{fig:overview_block}.

% \vspace{-.4cm}
\subsection{Experimental Results}

\begin{figure}[h]
\centering
 \includegraphics[width=0.27\textwidth]{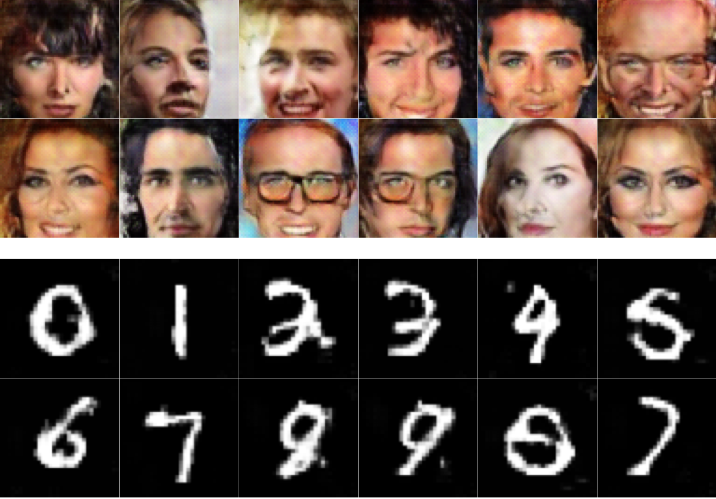}
\caption{Sample MNIST and CelebA images generated by the full precision DCNN.}
\label{fig:gen_result}
\end{figure}

\begin{figure}[h]
\centering
 \includegraphics[width=0.45\textwidth]{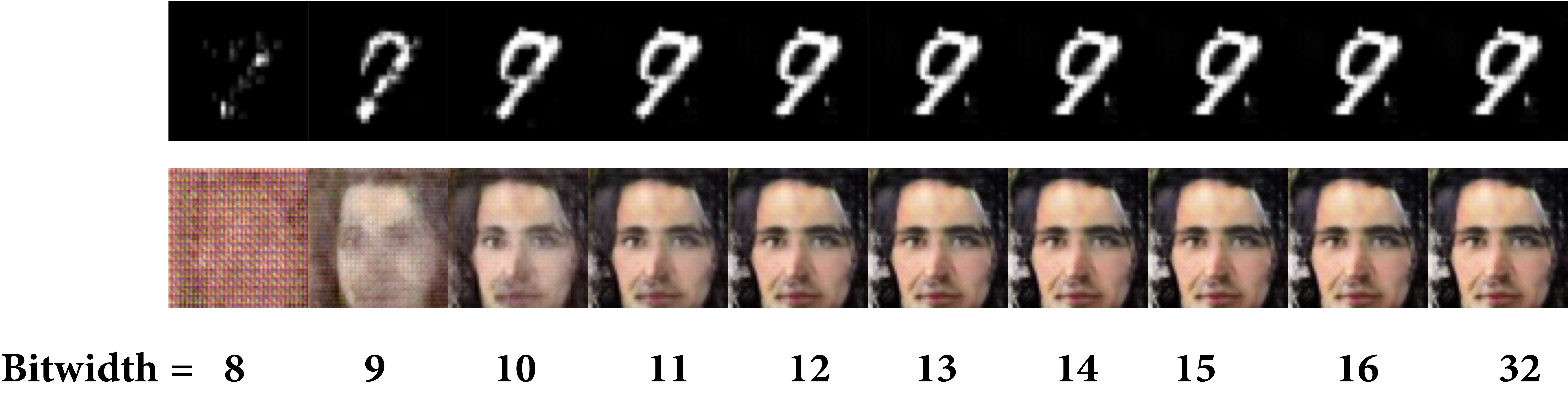}
\caption{Images generated by different bitwidth DCNNs.}
\label{fig:bit_width_result}
\end{figure}

Fig. \ref{fig:gen_result} shows some generated faces and digits from our trained DCNNs. Fig. \ref{fig:bit_width_result} shows the output of DCNNs under different bitwidths for the same input. Visually evaluating degradation of image quality is only feasible in the cases of extremely low bitwidth such as 8 bits. Our proposed methodology provides an analytical framework for quantifying the trade-off between image quality and implementation complexity over a range of bitwidths.

% Based on our statistical analysis, we choose the bitwidth to be 12. 

\begin{figure}[h] 
  \centering
  \includegraphics[width=0.4\textwidth]{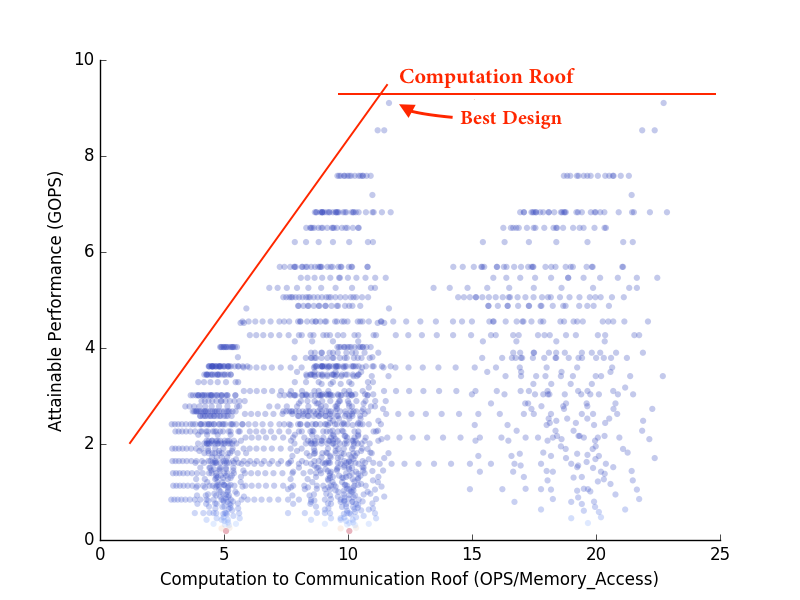}
  \caption{Design space Exploration for a layer with input 10x2x2 and output 64x4x4.}
  \label{fig:scatter}
\end{figure}

Fig. \ref{fig:scatter} shows all constraint-admissible design solutions for the first layer of our CelebA DCNN, where the best design is shown as located at the left corner of the roof. Table \ref{tab:util} shows the utilization rate after place and route, and we compare our DCNN performance with some existing CNN accelerators for reference in table \ref{tab:comp}. The performance can be further improved by implementing a ping-pong buffer in our system. 

% \vspace{-.5cm}

% Even though we don't have a ping pong buffer implemented in our system, 

\begin{table}[h]
\caption{FPGA Resource Utilization}
\begin{center}
\begin{tabular}{cccc}
\hline
DSP & LUT & FF & BRAM \\
\hline
95\% & 48\% & 29\% & 48\%\\
\hline
\end{tabular}
\end{center}
\label{tab:util}
\end{table}
% \vspace{-.2cm}
\begin{table}[h]
\caption{Comparison to previous implementations}
\begin{center}
\begin{tabular}{ccccccc}
\hline
& Chip & Precision & \#DSP & Freq & GOPS & GOPS/DSP \\
\hline
\cite{peemen2013memory} & VLX240T & Fixed & 768 & 150M & 17 & 0.022 \\
\hline
\cite{zhang2015optimizing} & VX485T & Float & 2800 & 100M & 61.62 & 0.022\\
\hline
Ours & 7Z020 & 12Fixed & 220 & 100M & 2.6 & \textcolor{red}{0.012}\\
\hline
\end{tabular}
\end{center}
\label{tab:comp}
\end{table}

%% file: content/6.Conclusion.tex
% \section{Related Work}
% \todo{this could be written very quickly, but need to discuss about which work to link with}
% \vspace{-.5cm}
\section{Conclusion}

In this work, we develop an FPGA-based deconvolution accelerator for deconvolutional neural networks and propose a three-step design methodology which first uses statistical analysis to find out the most cost-efficient bitwidth, then explore the design space with roofline model \cite{zhang2015optimizing} and use VLSI optimization methods to produce the final design. Finally, we implement our method on a Zynq-7000 FPGA and realize a performance density of 0.012 GOPs/DSP.

% \section{ACKNOWLEDGEMENTS}

% TODO

%% file: ms.bbl
\begin{thebibliography}{10}
\providecommand{\url}[1]{#1}
\csname url@samestyle\endcsname
\providecommand{\newblock}{\relax}
\providecommand{\bibinfo}[2]{#2}
\providecommand{\BIBentrySTDinterwordspacing}{\spaceskip=0pt\relax}
\providecommand{\BIBentryALTinterwordstretchfactor}{4}
\providecommand{\BIBentryALTinterwordspacing}{\spaceskip=\fontdimen2\font plus
\BIBentryALTinterwordstretchfactor\fontdimen3\font minus
  \fontdimen4\font\relax}
\providecommand{\BIBforeignlanguage}[2]{{%
\expandafter\ifx\csname l@#1\endcsname\relax
\typeout{** WARNING: IEEEtran.bst: No hyphenation pattern has been}%
\typeout{** loaded for the language `#1'. Using the pattern for}%
\typeout{** the default language instead.}%
\else
\language=\csname l@#1\endcsname
\fi
#2}}
\providecommand{\BIBdecl}{\relax}
\BIBdecl

\bibitem{schmidhuber2015deep}
J.~Schmidhuber, ``Deep learning in neural networks: An overview,'' \emph{Neural
  networks}, vol.~61, pp. 85--117, 2015.

\bibitem{lecun2015deep}
Y.~LeCun, Y.~Bengio, and G.~Hinton, ``Deep learning,'' \emph{Nature}, vol. 521,
  no. 7553, pp. 436--444, 2015.

\bibitem{Goodfellow-et-al-2016}
I.~Goodfellow, Y.~Bengio, and A.~Courville, \emph{Deep Learning}.\hskip 1em
  plus 0.5em minus 0.4em\relax MIT Press, 2016,
  \url{http://www.deeplearningbook.org}.

\bibitem{chakradhar2010dynamically}
S.~Chakradhar, M.~Sankaradas, V.~Jakkula, and S.~Cadambi, ``A dynamically
  configurable coprocessor for convolutional neural networks,'' in \emph{ACM
  SIGARCH Computer Architecture News}, vol.~38, no.~3.\hskip 1em plus 0.5em
  minus 0.4em\relax ACM, 2010, pp. 247--257.

\bibitem{chen2014dadiannao}
Y.~Chen, T.~Luo, S.~Liu, S.~Zhang, L.~He, J.~Wang, L.~Li, T.~Chen, Z.~Xu,
  N.~Sun \emph{et~al.}, ``Dadiannao: A machine-learning supercomputer,'' in
  \emph{Proceedings of the 47th Annual IEEE/ACM International Symposium on
  Microarchitecture}.\hskip 1em plus 0.5em minus 0.4em\relax IEEE Computer
  Society, 2014, pp. 609--622.

\bibitem{zhang2015optimizing}
C.~Zhang, P.~Li, G.~Sun, Y.~Guan, B.~Xiao, and J.~Cong, ``Optimizing fpga-based
  accelerator design for deep convolutional neural networks,'' in
  \emph{Proceedings of the 2015 ACM/SIGDA International Symposium on
  Field-Programmable Gate Arrays}.\hskip 1em plus 0.5em minus 0.4em\relax ACM,
  2015, pp. 161--170.

\bibitem{peemen2013memory}
M.~Peemen, A.~A. Setio, B.~Mesman, and H.~Corporaal, ``Memory-centric
  accelerator design for convolutional neural networks,'' in \emph{Computer
  Design (ICCD), 2013 IEEE 31st International Conference on}.\hskip 1em plus
  0.5em minus 0.4em\relax IEEE, 2013, pp. 13--19.

\bibitem{zeiler2010deconvolutional}
M.~D. Zeiler, D.~Krishnan, G.~W. Taylor, and R.~Fergus, ``Deconvolutional
  networks,'' in \emph{Computer Vision and Pattern Recognition (CVPR), 2010
  IEEE Conference on}.\hskip 1em plus 0.5em minus 0.4em\relax IEEE, 2010, pp.
  2528--2535.

\bibitem{3dgan}
J.~Wu, C.~Zhang, T.~Xue, W.~T. Freeman, and J.~B. Tenenbaum, ``Learning a
  probabilistic latent space of object shapes via 3d generative-adversarial
  modeling,'' in \emph{Advances in Neural Information Processing Systems},
  2016, pp. 82--90.

\bibitem{shi2016real}
W.~Shi, J.~Caballero, F.~Husz{\'a}r, J.~Totz, A.~P. Aitken, R.~Bishop,
  D.~Rueckert, and Z.~Wang, ``Real-time single image and video super-resolution
  using an efficient sub-pixel convolutional neural network,'' in
  \emph{Proceedings of the IEEE Conference on Computer Vision and Pattern
  Recognition}, 2016, pp. 1874--1883.

\bibitem{isola2016image}
P.~Isola, J.-Y. Zhu, T.~Zhou, and A.~A. Efros, ``Image-to-image translation
  with conditional adversarial networks,'' \emph{arXiv preprint
  arXiv:1611.07004}, 2016.

\bibitem{badrinarayanan2015segnet}
V.~Badrinarayanan, A.~Kendall, and R.~Cipolla, ``Segnet: A deep convolutional
  encoder-decoder architecture for image segmentation,'' \emph{arXiv preprint
  arXiv:1511.00561}, 2015.

\bibitem{wu2016quantized}
J.~Wu, C.~Leng, Y.~Wang, Q.~Hu, and J.~Cheng, ``Quantized convolutional neural
  networks for mobile devices,'' in \emph{Proceedings of the IEEE Conference on
  Computer Vision and Pattern Recognition}, 2016, pp. 4820--4828.

\bibitem{dundar1995effects}
G.~Dundar and K.~Rose, ``The effects of quantization on multilayer neural
  networks,'' \emph{IEEE Transactions on Neural Networks}, vol.~6, no.~6, pp.
  1446--1451, 1995.

\bibitem{goodfellow2014generative}
I.~Goodfellow, J.~Pouget-Abadie, M.~Mirza, B.~Xu, D.~Warde-Farley, S.~Ozair,
  A.~Courville, and Y.~Bengio, ``Generative adversarial nets,'' in
  \emph{Advances in neural information processing systems}, 2014, pp.
  2672--2680.

\bibitem{noh2015learning_seg}
H.~Noh, S.~Hong, and B.~Han, ``Learning deconvolution network for semantic
  segmentation,'' in \emph{Proceedings of the IEEE International Conference on
  Computer Vision}, 2015, pp. 1520--1528.

\bibitem{radford2015unsupervised}
A.~Radford, L.~Metz, and S.~Chintala, ``Unsupervised representation learning
  with deep convolutional generative adversarial networks,'' \emph{arXiv
  preprint arXiv:1511.06434}, 2015.

\bibitem{yu2015lsun}
F.~Yu, A.~Seff, Y.~Zhang, S.~Song, T.~Funkhouser, and J.~Xiao, ``Lsun:
  Construction of a large-scale image dataset using deep learning with humans
  in the loop,'' \emph{arXiv preprint arXiv:1506.03365}, 2015.

\bibitem{dumoulin2016guide}
V.~Dumoulin and F.~Visin, ``A guide to convolution arithmetic for deep
  learning,'' \emph{arXiv preprint arXiv:1603.07285}, 2016.

\bibitem{bounliphone2015test_rmmd}
W.~Bounliphone, E.~Belilovsky, M.~B. Blaschko, I.~Antonoglou, and A.~Gretton,
  ``A test of relative similarity for model selection in generative models,''
  \emph{arXiv preprint arXiv:1511.04581}, 2015.

\bibitem{gretton2012kernelmmd}
A.~Gretton, K.~M. Borgwardt, M.~J. Rasch, B.~Sch{\"o}lkopf, and A.~Smola, ``A
  kernel two-sample test,'' \emph{Journal of Machine Learning Research},
  vol.~13, no. Mar, pp. 723--773, 2012.

\bibitem{xilinx:ug479}
\emph{7 Series DSP48E1 Slice}, XILINX INC, 9 2016, rev. 1.9.

\bibitem{Coussy:2008:HSA:1457713_hls}
P.~Coussy and A.~Morawiec, \emph{High-Level Synthesis: From Algorithm to
  Digital Circuit}, 1st~ed.\hskip 1em plus 0.5em minus 0.4em\relax Springer
  Publishing Company, Incorporated, 2008.

\bibitem{neopane2016nonparametric}
O.~Neopane, S.~Das, E.~Arias-Castro, and K.~Kreutz-Delgado, ``A nonparametric
  framework for quantifying generative inference on neuromorphic systems,'' in
  \emph{Circuits and Systems (ISCAS), 2016 IEEE International Symposium
  on}.\hskip 1em plus 0.5em minus 0.4em\relax IEEE, 2016, pp. 1346--1349.

\bibitem{liu2015faceattributes}
Z.~Liu, P.~Luo, X.~Wang, and X.~Tang, ``Deep learning face attributes in the
  wild,'' in \emph{Proceedings of International Conference on Computer Vision
  (ICCV)}, 2015.

\end{thebibliography}
